\documentclass[review]{elsarticle}
\usepackage[hidelinks]{hyperref}
\usepackage{footnote}
\makesavenoteenv{tabular}
\usepackage{soul}
\usepackage{caption}
\usepackage{subcaption}
\usepackage[hmargin=2.5cm]{geometry}

\journal{Computers in Biology and Medicine}

\begin{document}

\begin{frontmatter}

\title{Validating an SVM-based neonatal seizure detection algorithm for generalizability, non-inferiority and clinical efficacy}

\fntext[myfootnote]{Department of Neuroscience and Biomedical Engineering
Aalto University School of Science
P.O. Box 12200
FI-00076 AALTO, Finland}

\author[add1,add2,add3,add4]{Karoliina T. Tapani}
\ead{karoliina.tapani@aalto.fi}
\author[add1,add5]{Päivi Nevalainen}
\author[add1]{Sampsa Vanhatalo}
\author[add3]{Nathan J. Stevenson}
\address[add1]{BABA Center, Helsinki University Hospital, University of Helsinki, Finland}
\address[add2]{Department of Neuroscience and Biomedical Engineering, Aalto University, Espoo, Finland$^1$}
\address[add3]{Brain Modelling Group, QIMR Berghofer Medical Research Institute, Australia}
\address[add4]{Department of Medical Physics, Kymenlaakso Central Hospital, Kotka, Finland}
\address[add5]{Epilepsia Helsinki and Department of Clinical Neurophysiology, HUS diagnostic center, Helsinki University Hospital and University of Helsinki, Finland}



\begin{abstract}

Neonatal seizure detection algorithms (SDA) are approaching the benchmark of human expert annotation. Measures of algorithm generalizability and non-inferiority as well as measures of clinical efficacy are needed to assess the full scope of neonatal SDA performance. We validated our neonatal SDA on an independent data set of 28 neonates. Generalizability was tested by comparing the performance of the original training set (cross-validation) to its performance on the validation set. Non-inferiority was tested by assessing inter-observer agreement between combinations of SDA and two human expert annotations. Clinical efficacy was tested by comparing how the SDA and human experts quantified seizure burden and identified clinically significant periods of seizure activity in the EEG. Algorithm performance was consistent between training and validation sets with no significant worsening in AUC ($p>0.05$, $n =28$). SDA output was inferior to the annotation of the human expert, however, re-training with an increased diversity of data resulted in non-inferior performance ($\Delta\kappa$=0.077, 95\% CI: -0.002-0.232, $n=18$). The SDA assessment of seizure burden had an accuracy ranging from 89-93\%, and 87\% for identifying periods of clinical interest. The proposed SDA is approaching human equivalence and provides a clinically relevant interpretation of the EEG.

\end{abstract}


\end{frontmatter}


\section{Introduction}

Clinical suspicion of neonatal seizures or an identified risk of seizures due to e.g. hypoxic–ischaemic encephalopathy (HIE), meningitis or stroke are key indications for electroencephalography (EEG) monitoring in the neonatal intensive care units  \cite{rennie2007}. To guide interventions and inform prognosis, reliable and accurate detection of seizures is essential \cite{bjorkman2010seizures,miller2002seizure}. Recent guidelines insist that neonatal seizures must have an EEG correlate as clinical recognition of seizures is confounded by a range of factors \cite{pressler2021ilae}. Neonatal EEG monitoring generates a considerable amount of data and its interpretation is a challenge that has led to the development of several automated seizure detection algorithms (SDA)  \cite{liu1992detection,gotman1997automatic,celka2002computer,smit2004neonatal,navakatikyan2006seizure,deburchgraeve2008automated,mitra2009multi,temko2011eeg,ansari2016improved,o2017neonatal,tapani2019,ansari2019neonatal,o2020neonatal,isaev2020attention}.

While the performance of many of these algorithms has thoroughly been assessed within the training data, few of them have been validated on an independent data set \citep{gotman1997evaluation,smit2004neonatal,lawrence2009pilot,cherian2011validation,mathieson2016validation, o2020neonatal}. This step is essential as it ensures that SDAs built on relatively small data sets of neonatal EEG generalize to the larger population of neonatal EEG recordings. While generalizability is important, it alone does not suffice to define SDA performance. An SDA must also be accurate and clinically useful. 
The complete assessment of SDAs, therefore, must address four important questions: 1) does the SDA performance generalize?; 2) does the SDA generate an annotation that is non-inferior to the human expert, taking into account ambiguity in the annotations of human experts?; 3) does the annotation of the SDA accurately reflect the seizure burden experienced by a neonate over a clinically relevant period of time (e.g. birth to 72-96 h post-natal age)?; 4) will the implementation of the SDA improve clinical practice?

The SDA developed by Temko et al. (2011) has the most comprehensive assessment to date \cite{temko2011performance}. Promising early performance was validated on an independent data set and the annotation of the SDA was shown to provide an accurate reflection of seizure burden \cite{mathieson2016validation}. Non-inferiority of the SDA annotation to the annotation of the human expert was, however, not comprehensively answered and the performance of the SDA when clinically trialled was underwhelming \cite{pavel2020machine}. Recent advances in SDAs and subsequent improvements in accuracy, hold out hope that modern SDAs will soon achieve a level of performance that will, ultimately, improve clinical practice.

In this paper we 1) assess the generalizability of our previously developed SDA on an unseen validation data set; 2) determine if the SDA generates an annotation that is non-inferior to the human expert; and 3) determine if the annotation of the SDA accurately estimates the seizure burden on a neonate. While we do not have access to the implementation of the SDA of Temko et al. (2011), we will use our implementation of their feature set, trained on our data, as a benchmark.

\section{Materials and Methods}

This study utilizes two cohorts of EEG recordings. The first data set (training data set) is used to develop and train the neonatal SDA. The second data set (validation data set) is used to validate the SDA. To test the ability of the SDA to generalize to unseen data, several performance measures evaluating differences between the SDA and human annotation of the EEG, were compared between the training and validation data sets. The smaller the difference in these performance measures the more generalized the SDA. The non-inferiority of the SDA annotation compared to the human annotation of the validation set was examined using measures of inter-observer agreement (IOA). Clinically relevant measures of seizure burden were compared between the SDA and human expert annotations to assess the clinical efficacy of the algorithm. Finally, the validation data set was split in two, with the first third ($n=10$) added to the training set and the latter two thirds ($n=18$) saved as a final test set. This process helps determining if more diverse training data allows improving SDA generalizability or non-inferiority. 


\subsection{EEG acquisition and annotation}

\textit{Training data set}: This data set consists of EEG recordings from a cohort of 79 neonates with a variety of aetiologies. These neonates were recorded with a short duration paradigm and a whole scalp, 10-20 EEG system between 2010 and 2014. The EEG was recorded using a Nicolet One EEG monitor (Nicolet One, Natus Medical Incorporated, USA) with 19 scalp electrodes placed on positions Fp1, Fp2, F3, F4, F7, F8, T3, T3, C3, C4, T5, T6, P3, P4, O1, O2, Fz, Cz, Pz with sampling rate of 256Hz. The median duration of each recording was 74 min (IQR: 64 to 96 min). A bipolar derivation (double banana) was annotated by three clinical experts and used to train the SDA. The data set contains 460 seizures from 39 neonates (in consensus). More details can be found in \cite{stevenson2019dataset}.

\textit{Validation data set}: This data set consists of EEG recordings from a cohort of 28 neonates with stroke or HIE (Table \ref{tab:demo}). This validation set is a subset of the cohort first studied in \citep{nevalainen2019bedside}.

\begin{table}[ht]
\caption{\label{tab:demo} Table of patient characteristics for the validation data set ($n=28$). Data shown as $n$ (\%), or mean (standard deviation).} 
\begin{center}
\begin{tabular}{ll}
\hline
Patient characteristics ($n=28$)  &            \\ \hline
Gestation age (weeks)             & $39.2 (\pm 2)$   \\
Birth weight (g)                  & $3300 (\pm 550)$ \\
pH                                & $7.2 (\pm 0.11)$ \\ \hline \hline
Etiologies                        &            \\ \hline
Stroke                            & 25 (89\%)  \\
Hypoxic-ischemic encephalopathy   & 3 (11\%)   \\ \hline 
\end{tabular}
\end{center}
\end{table}

These neonates were recorded with long duration brain monitoring between 2011 and 2016. The signal was acquired at 250Hz using a Nicolet One EEG monitor (Nicolet One, Natus Medical Incorporated, USA) and four needle electrodes placed on positions F3, F4, P3, P4.  The total length of all recordings was 86 days 16 hours (median 77h 50min, IQR: 48h 20min - 94h 35min). A three channel, bipolar derivation (F3-P3, F4-P4, P3-P4) was annotated by two clinical experts (PN, SV: both employed at Helsinki University Hospital) and processed by the SDA. The start and stop times of each seizure were noted. For annotations, the internationally approved neonatal seizure definition by Clancy was used \cite{clancy1987exact}. Clinical experts were blinded to each others annotations.

The training and validation sets were both collected in the neonatal intensive care unit (NICU) of the Children's Hospital, Helsinki University Hospital, Finland. The EEG was recorded as part of the standard care in our NICU. The use of the training recordings were approved by the Ethics Committee of the Helsinki University Children’s Hospital, Finland. The Institutional Research Review Board at Helsinki Children’s Hospital approved the extraction and collation of the validation data set from hospital records and waived consent due to the study’s retrospective and observational nature.

\subsection{SDA}

This SDA is based on a 'bag of features' calculated on 16s epochs of EEG. The combination of features to form a decision statistic was performed with a support vector machine (SVM) \cite{tapani2019}. We used 22 summary EEG measures/features to represent a 16 s epoch of EEG. One additional feature to the previously published algorithm was maximum amplitude, which we used to discard epochs with high amplitude. To improve the robustness of the SDA, we incorporated an outlier detection system into our SDA. This was due to the relatively small size of our database and the expected difference in EEG quality between high density, short duration training data and low density, long duration validation data. Therefore we used a modified version of our previous system that was first proposed in \cite{tapani2019, stevenson2019hybrid}. Post-processing of the initial SDA output included eliminating outliers (we also eliminated EEG epochs detected as 'bad electrode' by the EEG monitor), applying a temporal moving average to each channel, taking the maximum value across channels and thresholding to form a binary (seizure/no seizure) decision. The binary decision was then extended in time using a collar. The parameters for outlier determination and seizure detection were optimized on training data by maximizing the agreement between the SDA and human annotation.

\subsubsection{Outlier detection}

Outlier detection identifies periods of the EEG recording that fall outside its experience; in our case the span of the feature set calculated on the training data. These methods are important to incorporate into clinical decision algorithms as the full variety of EEG patterns may never be captured in training data sets and when the detection of the rarer, clinically important class (seizure) is critical. The implementation of outlier detection, therefore, ensures that unusual EEG patterns (typically associated with artefact or rare neurological conditions) which fall outside the span of training data are dealt with differently. For our SDA, we adhere to the principle of \textit{primum non nocere} and annotate 'no seizure' when an outlier is detected.

\begin{figure}[ht]
  \centering
  \includegraphics[width=0.9\linewidth]{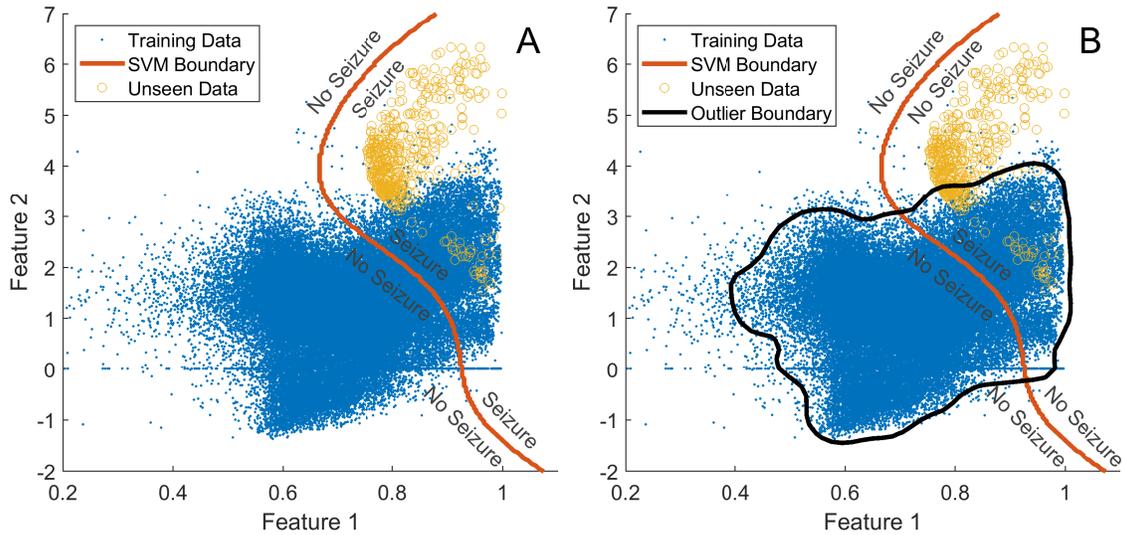}  
\caption{The effect of outlier detection on SDA annotation. The blue dots denote most of the training samples in the database. A) The original SVM based SDA. The thick red line is the SVM decision boundary for determining ‘seizure’ or ‘no seizure’ and the yellow circles represent incoming, unseen data (potential false detections). B) SVM based SDA with outlier detection. The black boundary is based on kNN distances of the training data; the updated decision boundary reduces the size of the feature space assigned to 'seizure'.}
\label{fig:2D}
\end{figure}

The effect of outlier detection on seizure annotation is shown in Figure \ref{fig:2D}. The example is based on a 2 feature SVM. In \ref{fig:2D}A, the original SVM shows the separation between the two classes, 'seizure' and 'no seizure'. In \ref{fig:2D}B the black outlier boundary is included and is based on distances of the 3 nearest neighbours within the training data. Any data falling outside the decision boundary is labelled as 'no seizure' resulting in a smaller decision space for 'seizure'.

As not all features in our original feature set are best suited for outlier detection, we use both general feature space distances, $k$-nearest neighbors (kNN) search, and feature specific (EEG amplitude) distances. If an incoming data point is either greater than a threshold Euclidean distance from the k-nearest neighbours in the training data set or greater than a threshold EEG amplitude, then an outlier is detected and no seizure annotated. 

We used the knnsearch function in Matlab (\cite{friedman1977kNN}), which outputs ascending Euclidean distances of an incoming data point to an array of training data points. In the case of our SVM-based detector, this data point is represented by a vector of features computed on a single epoch of EEG. The number of distances measured, that is the number of neighbors, is defined by parameter $k$. 

We complemented our thresholding of feature space distances with a stricter, separate threshold on EEG amplitude. The number of neighbours, distance threshold for the kNN search, EEG amplitude threshold, decision collar and decision threshold were, therefore, optimized on the training data set. 


\subsection{Analysis of algorithm performance}


The most commonly used measure in SDA performance assessment is the area under the curve of the receiver operator characteristics (AUC), a second by second comparison of the algorithm annotation to the human annotation. It combines two important measures, true positive rate (or sensitivity) and false detection rate (or specificity) on multiple detection thresholds. These measures can be defined with the numbers of true positives/seizure (TP), true negatives/non-seizure (TN), false positives (FP) and false negatives (FN). Sensitivity is defined by TP/(TP+FN) and specificity by TN/(TN+FP). In addition to these temporal measures we assessed seizure detection rate (SDR), that is the number of correctly detected seizure events, and false detections per hour (FD/h), which is the number of falsely detected seizure events divided by the duration of the recording in hours. These measures can be supplemented with measures of inter-observer agreement (IOA) such as the Kappa family of statistics; we use Cohen's kappa as we only compare between two annotations at a time \citep{cohen1960coefficient}.

We calculated performance measures on individual neonates and also on a single, concatenated, consensus of human expert annotation for comparison. As each data set contained neonates with no seizure, a concatenated annotation avoids complications of per neonate assessments (sensitivity/AUC and SDR cannot be calculated on neonates who do not have seizure). It also provides a more useful assessment of event based measures as several recordings in the training data set had limited events on relatively short recordings resulting in poor quantization per neonate.

\subsubsection{Implementation of Outlier Detection}

The cross-validation results of the original SDA and SDA with outlier detection were compared using statistical testing. Differences between AUC, sensitivity and specificity calculated on a per neonate basis were tested using Wilcoxon signed rank test (for paired data) due to non-normal distribution of these values across training and validation cohorts. AUC and sensitivity can only be calculated on neonates with seizure, while specificity was calculated on all neonates in the training and validation cohorts. All tests were two-sided and the level of significance was 0.05. 

Analysis of concatenated recordings was also performed using bootstrapped statistics. In this case, a concatenated annotation from each data set was generated using a random sampling of neonate annotations (sampling with replacement). The difference between performance measures were calculated on each re-sampled annotation. The difference in performance measures was calculated for 1000 re-samplings and used to estimate a 95\% confidence interval of differences. If the difference spanned zero, there was no significant difference in performance between data sets. This statistical method was used to compare performance where applicable (we cannot compare bootstrapped distributions between different data sets).

As outlier parameters were optimized using IOA between the algorithm and human experts, we used Cohen's kappa as the primary measure of performance differences between the SDA, with and without, outlier detection. 

\subsubsection{Generalizability of the SDA}

We measured generalization by comparing perfmance measures between cross-validation assessments of the training dataset and the validation dataset. If there was no statistically significant difference in these measures between the training and validation sets then we claim that the SDA \textit{generalizes} to a larger population of neonates. The primary measure of generalizability was statistical testing of AUC. 

Differences between AUC, sensitivity and specificity calculated on a per neonate basis were tested similarly to the previous section, with the exception that a Mann-Whitney U tests (for unpaired data) was used. If the difference in AUCs is not significant, the SDA generalizes on unseen data.

\subsubsection{Non-inferiority to human annotation}

The limitation of AUC as a measure of performance is that it does not inherently determine if the SDA annotation of the EEG is 'good enough'. This is particularly valid in instances where there is ambiguity in the 'gold standard' used to calculate the AUC such as the manual annotation of the EEG by human experts. We, therefore, determined the non-inferiority of the SDA to the annotations of the human expert using bootstrapped measures of IOA. Each expert and the SDA were compared to each other using Cohen's kappa statistic. The difference between the kappa of SDA/human expert pairs and human/human pairs was estimated using a bootstrap (1000 iterations). If the difference spanned zero then the SDA/human kappa was assumed to be equivalent/non-inferior to the human/human kappa. This analysis uses the complete annotation of each human expert and avoids the requirement of consensus annotations.

We also investigated SDA errors at the annotation and feature level. At the annotation level, we compared general summary measures of the annotation such as the total seizure burden, seizure number and seizure duration. At the feature level, we determined if some features were systematically different between the training and validation data sets. 


\subsubsection{Clinical efficacy}

While non-inferiority provides an effective means to compare the annotation of an SDA to the annotation of human experts, it does not evaluate the practical clinical impact of any discrepancies. To this end, we determined the clinical efficacy of SDA by comparing clinically relevant interpretations of SDA and human annotations of the EEG. The first set of measures compares a continuous assessment of seizure burden. This measure is considered clinically important as it provides a clear quantitative target for treatment.

While some clinicians are aggressively treating any amount of seizures with a variety of anti-epileptic medications \cite{rennie2007,van2013treatment,wickstrom2013differing}, there is also substantial variance among clinicians with respect to the level of seizure burden that is considered to prompt treatment. We compared hourly assessments of seizure burden using the correlation coefficient and a diagnostic binary decision of periods of clinical targets and periods of no clinical target in an EEG recording. We define a clinical target as a two-hour period with two 30s seizures or 3 minutes of accumulated seizure \cite{pressler2021ilae}. We then compare the number of clinical targets defined by the human annotation to the number defined by SDA annotation using sensitivity and specificity.

We also evaluate the prognostic utility of the SDA. Recent work has shown that high total seizure burden ($>$ 45 minutes) and high maximum hourly seizure burden ($>$ 13 minutes per hour) in a long duration recording of EEG are associated with poor neurodevelopmental outcome in neonates with HIE \cite{kharoshankaya2016seizure}. Note, that a high hourly seizure burden is typically used to define status epilepticus ($>$30 minutes per hour). We use 13 minutes per hour as an alternate definition of high seizure burden due to its association with poor developmental outcome \cite{kharoshankaya2016seizure}. We compare the SDA annotation to the human consensus annotation for the prediction of high or low seizure burden in each neonate using measures of sensitivity and specificity. We also show how differences in human interpretations affect these measures by averaging sensitivity and specificity between experts, using expert 1 as the 'gold standard' and using expert 2 as the 'gold standard'. 

\subsection{Re-training}

Finally, we investigated the incorporation of additional data from the validation set into the training set on SDA performance. We extracted seizure and non-seizure epochs from the first 10 neonates in the validation data set. The epochs for each class were randomly sampled from consensus segments, so that the size of this new training set was comparable to the size of the original training set. The non-seizure epochs were randomly sampled from the second half of the recordings only, as the majority of false detections occurred in this time period (see Results). To increase diversity in the training data set, not the total number of samples, every second sample from the new and from the original training data was used to re-train the SVM. Generalizability and non-inferiority measures were calculated on the re-trained SVM to determine the effectiveness of added diversity in the training data. 

\section{Results}

\subsection{Implementation of outlier detection and generalization of the SDA}

The incorporation of an outlier detection significantly improved the agreement between the SDA output and the consensus annotation of the human expert (Table \ref{tab:gen}). Slight, but significant reductions in per infant AUC were offset by significant improvements in specificity and false detections per hour.

The SDA had consistent performance between training and validation sets with our primary measure for generalizability, the AUC, suggesting that the algorithm generalizes to unseen data (Table \ref{tab:gen}). That is, there was no significant reduction in performance between the cross-validated training data set and the validation data. The decrease in sensitivity was significant, while there was no statistical difference for specificity. 

The proposed SDA outperformed an SDA using the Temko feature set (Table \ref{tab:gen}). 

\begin{table}[ht]
\caption{\label{tab:gen} The performance of a neonatal SDA with outlier detection. The neonatal SDA with outlier detection was applied to the original dataset (10-fold cross-validation) and compared to the original SDA without outlier detection. The neonatal SDA was also applied to an independent validation set and compared to an SDA based on the Temko feature set (Temko FS). Median (IQR) are presented for AUC, sensitivity (Sens) and specificity (Spec). The value with 95\%CIs in brackets are presented for cAUC, cSDR, cFD/h and cKappa (estimated on concatenated recordings). $^a$ and $^b$ denote a significant increase and decrease, respectively, between the original SDA and the SDA with outlier detection. $^c$ denotes a significant increase in SDA performance between the training and validation data sets. $^d$ denotes significant decrease in performance between proposed SDA and SDA based on Temko features. Primary measures of performance are shown in bold. $n_t$ is the total number of neonates in each dataset and $n_s$ is the number of neonates with consensus seizure in each dataset.} 
\begin{center}
\begin{tabular}{| c  c c | c | c |}
\hline
	& Original Training  & Original Training & Validation & Validation \\ 
	& (10-fold CV) & (10-fold; outlier detection) & (outlier detection) & (Temko FS) \\
	& $n_t = 79, n_s = 39$ & $n_t = 79, n_s = 39$ & $n_t = 28, n_s = 24$ & $n_t = 28, n_s = 24$ \\ \hline
	AUC & 0.992 (0.938-0.998) &   0.986 (0.908-0.995)$^b$ & \textbf{0.960 (0.896-0.980)} & 0.918 (0.823-0.961)$^d$ \\ 
    Sens & 0.823 (0.341-0.975) & 0.784 (0.363-0.958)  & 0.605 (0.218-0.753)$^c$ & 0.385 (0.061-0.674)$^d$ \\
	Spec & 0.993 (0.979-1) &   0.996 (0.982-1)$^a$ & 0.998 (0.997-0.999) & 0.998 (0.994-0.999)$^d$\\ \hline
	
	cAUC & 0.956 (0.931-0.979) & 0.955 (0.928-0.977) & 0.967 (0.953-0.974) & 0.949 (0.931-0.961)$^d$ \\
	cSDR & 0.750 (0.582-0.851) &  0.685 (0.531-0.791)  & 0.761 (0.631-0.841) & 0.657 (0.522-0.750)$^d$ \\ 
	cFD/h & 2.290 (1.312-3.897) & 1.967 (0.964-3.478)$^b$ & 0.357 (0.167-0.690) & 0.302 (0.135-0.513) \\
	cKappa & 0.668 (0.464-0.749) & \textbf{0.672 (0.474-0.764)$^a$ } & 0.729 (0.656-0.789) & 0.623 (0.533-0.694)$^d$ \\
	 \hline	
\end{tabular}
\end{center}
\end{table}


\subsection{Non-inferiority to human annotation}

\begin{figure}
     \centering
     \begin{subfigure}[b]{0.75\textwidth}
        \centering
         \includegraphics[width=\textwidth]{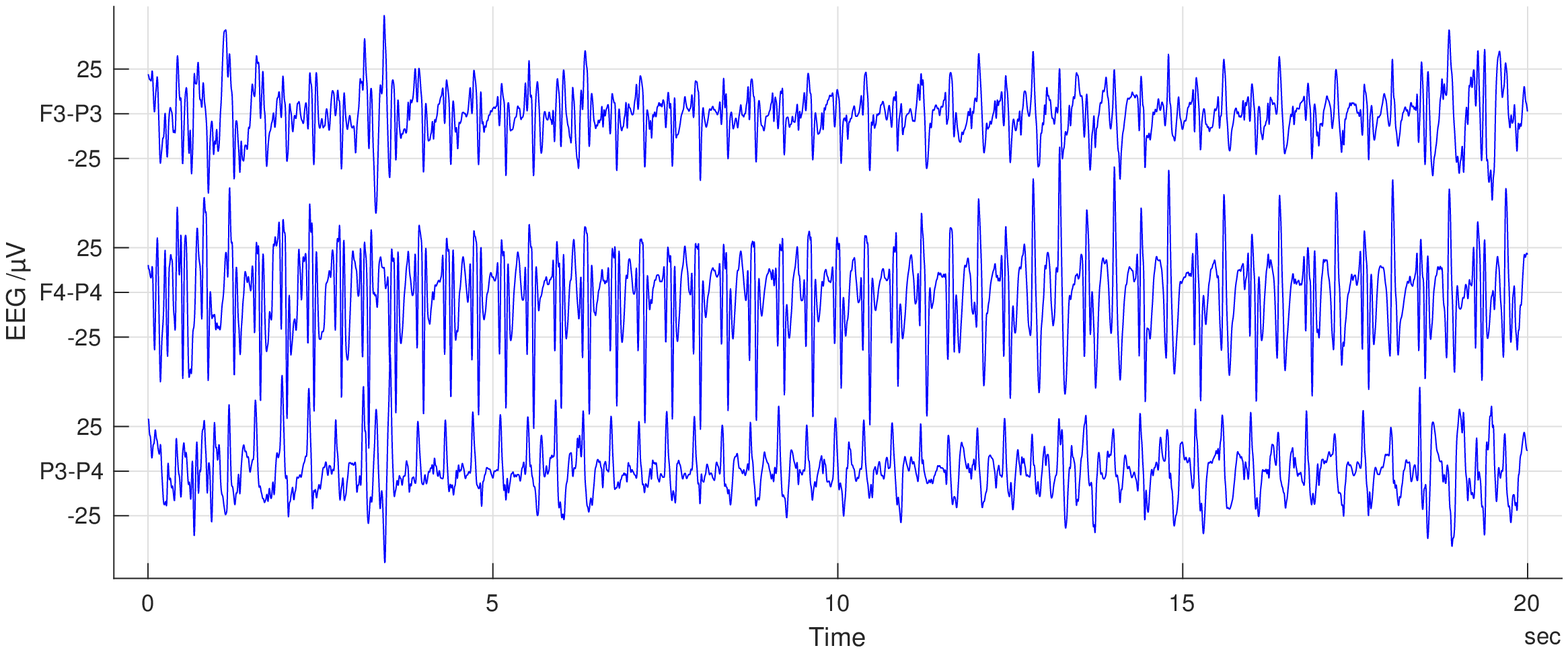}
         \caption{True detection on patient 2. The seizure is on all channels, but most prominent on channel F4-P4.}
     \end{subfigure}
     \begin{subfigure}[b]{0.75\textwidth}
     \centering
         \includegraphics[width=\textwidth]{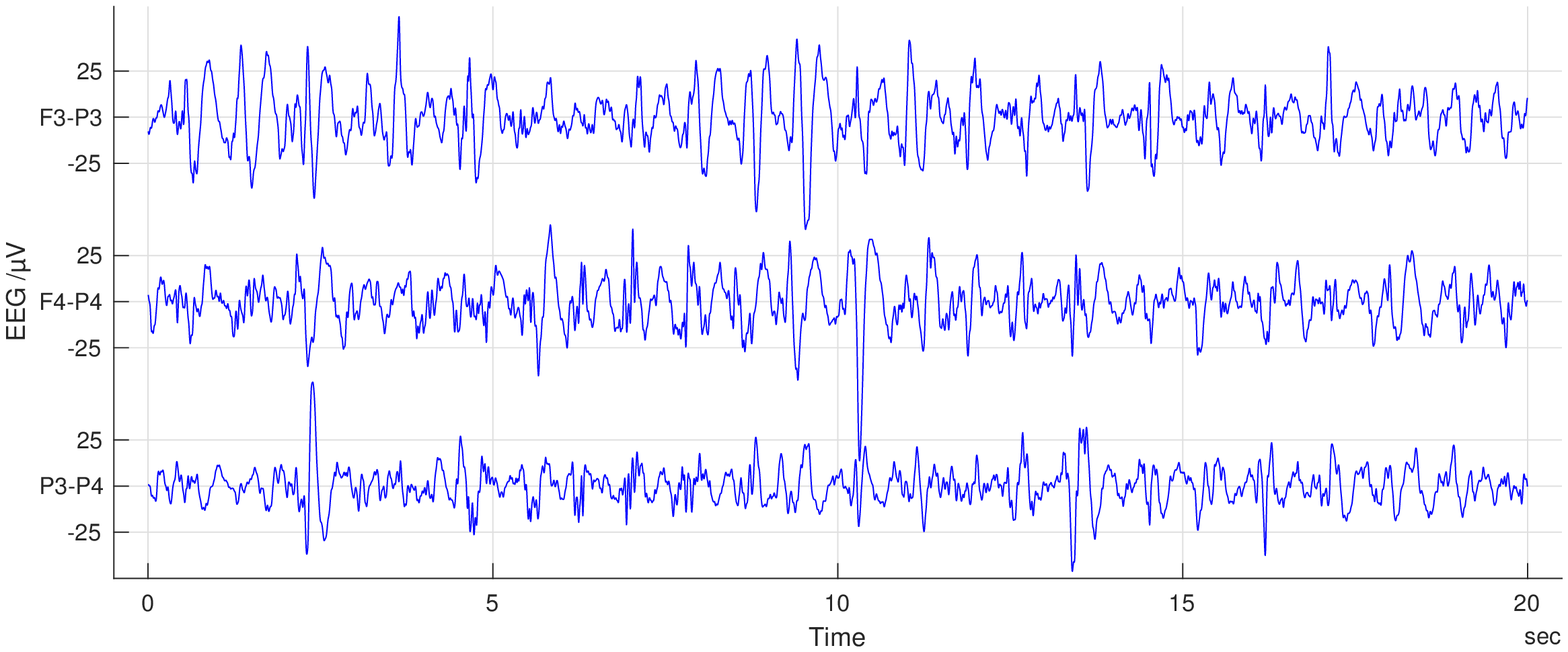}
         \caption{False detection on patient 34. Channel F3-P3 raises the false alarm. }
     \end{subfigure}
          \hfill
     \begin{subfigure}[c]{0.75\textwidth}
         \centering
         \includegraphics[width=\textwidth]{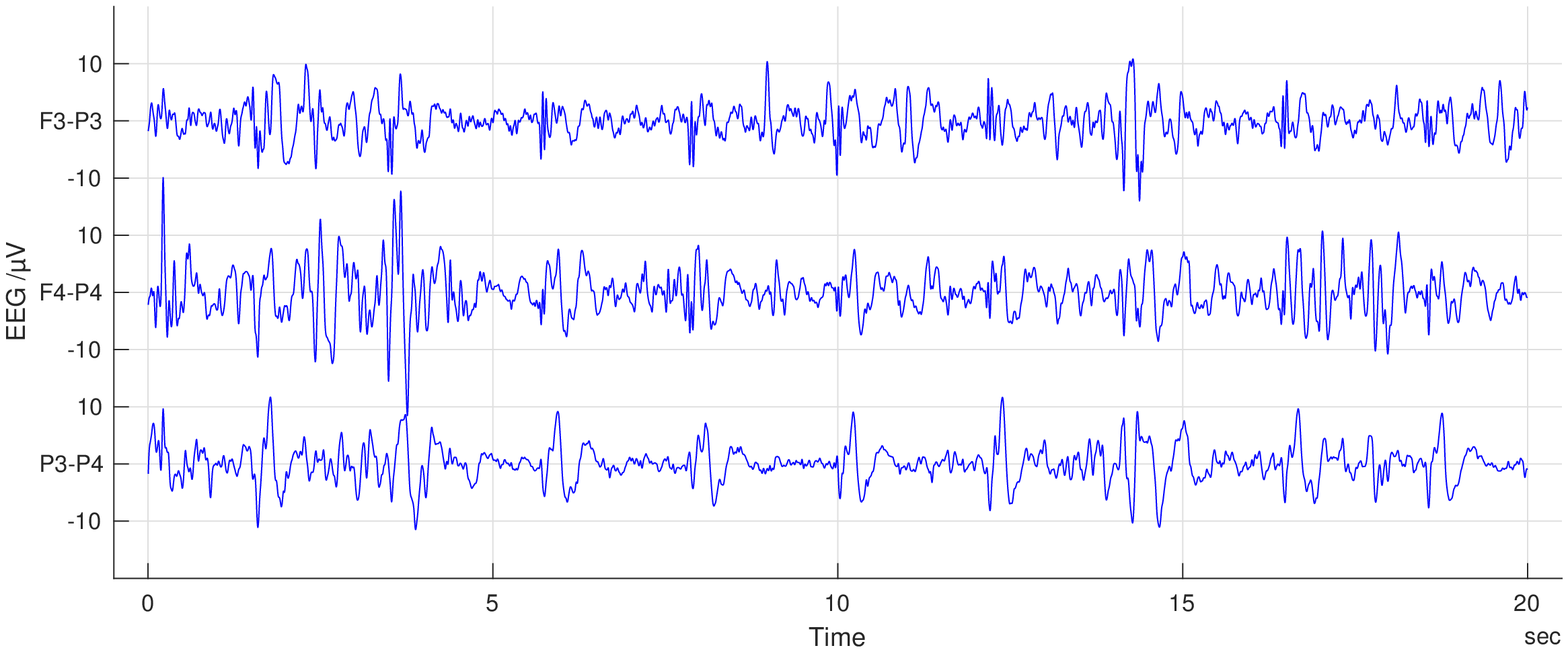}
         \caption{Missed seizure on patient 1. Seizure is on all channels, but predominantly on channel P3-P4.}
     \end{subfigure}
        \caption{Examples of a true detection, a false detection and a missed seizure.}
        \label{fig:fd_fn}
\end{figure}

The SDA provides an annotation of the EEG that is slightly inferior to human experts (Table \ref{tab:ioa}; $\Delta\kappa$ fails to span 0). Example EEG segments that were correctly detected as seizure, missed or falsely labelled as seizure are shown in Figure \ref{fig:fd_fn}. 

\begin{table}[ht]
\caption{\label{tab:ioa} Inter-observer agreement between human and algorithm based annotations of seizure/non-seizure in the validation set of EEG. E1 is human expert 1, E2 is human expert 2, Consensus is the consensus annotation of human experts, $\Delta\kappa$ is difference in IOA between a single human expert and the SDA and the IOA between human experts (results are presented as median, 95\% confidence interval), FS is feature set.}
\begin{center}
\begin{tabular}{lccc}
\hline \hline
Inter-observer Agreement     & E1                   & E2                   & Consensus        \\
\hline
SDA (proposed FS)            & 0.591                & 0.703                & 0.729            \\
SDA (Temko FS)            & 0.511                & 0.600                & 0.623           \\
E1             & -                    & 0.792                & 0.805            \\
E2             & -                    & -                    & 0.982            \\
\hline
SDA $\Delta\kappa$ (proposed FS) & \textbf{0.197 (0.149-0.269)}  & \textbf{0.085 (0.011-0.176)}  &  - \\
SDA $\Delta\kappa$ (Temko FS) & \textbf{0.277 (0.220-0.364)}  & \textbf{0.187 (0.113-0.289)}  &  -\\
\hline \hline
\end{tabular}
\end{center}
\end{table}




At the feature level, the major differences between training and validation data sets were between the smoothed nonlinear energy operator feature within EEG epochs without seizure (training data: median [IQR] 39.2 [15.2-74.8], validation data: median [IQR] 69.9 [43.5-111.4]; values are not normalized). Increases in the nonlinear energy imply increases in high frequency energy in seizure signals.

\begin{figure}[htp!]
    \centering
    \includegraphics[width=0.75\textwidth]{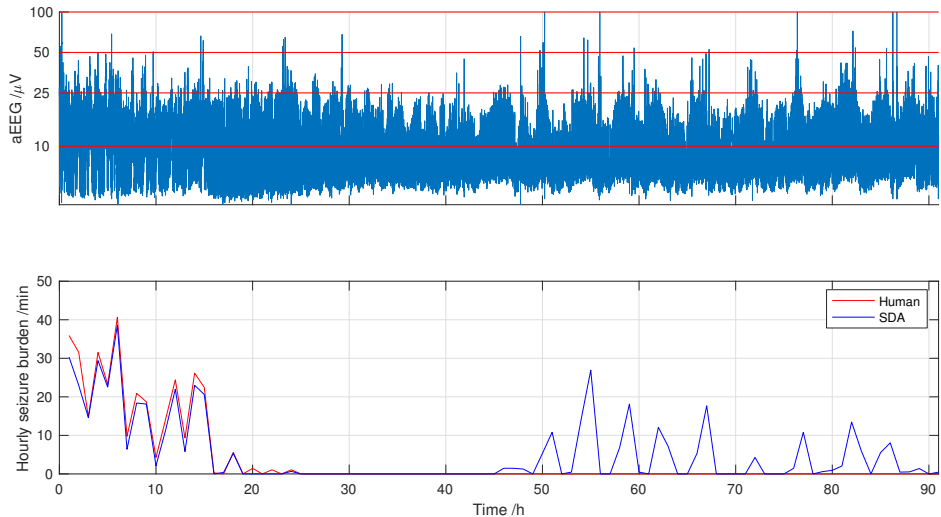}
    \caption{An example aEEG and seizure burden time course. Note the appearance of sleep-wake-cycling after 50 h. This will introduce normal EEG patterns such as mixed frequency activity, high voltage synchronous activity and trace alternant, that will not be prevalent in the original training data as those recordings were typically taken earlier in the seizure burden time course.}
    \label{fig:sbx}
\end{figure}

At the annotation level, the SDA has a noticeable reduction in seizure detection rate and, in general, it generates an excessive number of seizures compared to the consensus seizure annotation. This is in the context of an SDA annotation that is, nevertheless, more conservative and diffuse with its annotation; the SDA has lower estimates of seizure burden measures (maximum and total, Table \ref{tab:stats}). The SDA also tends to generate more false detections during the second half of recordings compared to the first. A representative example is shown in Figure \ref{fig:sbx}. In general, the median false detection rate was 0.13 per hour (IQR: 0.03-0.17) during the first half of the recording and 0.20 per hour (IQR: 0.06-0.42) during the second half of the recording. The second half of recordings is more likely to contain EEG patterns associated with recovery and, therefore, more normal EEG such as sleep states. These patterns may not be prevalent in the training data. 

\begin{table}[htp!]
\caption{\label{tab:stats} Seizure summary statistics of seizures annotated by SDAs and the consensus of human experts. Results are presented as median (IQR) summarized across neonates with seizures. Seizure duration is presented as median (IQR) over all annotated seizures. $^1$ 24 neonates had seizures annotated by both human consensus and the SDA (proposed FS) and $^2$ 23 by consensus and SDA (Temko FS).}
\begin{center}
\begin{tabular}{lccc}
\hline \hline 
Seizure statistics                 & Consensus & SDA (proposed FS) & SDA (Temko FS) \\
\hline
Seizure patients            & 24           & 28$^1$       & 27$^2$ \\
Seizure number              & 33  (11-68)  & 29 (17-119) & 14 (8-120) \\
\textbf{Seizures (total)}   & \textbf{1362}& \textbf{2313}& \textbf{1749}\\
Seizure duration (s)        & 96 (55-167)  & 44  (32-84)  &  64 (52-104) \\
Burden (max; min/h)         & 13 (3-29)    & 8 (3-20)    & 9 (2-21)\\
Burden (total; min)         & 42 (11-163) & 28 (10-138) & 24 (8-201)\\
\hline \hline
\end{tabular}
\end{center}
\end{table}

\subsection{Clinical efficacy}

\begin{figure}[htp!]
     \centering
         \includegraphics[width=\textwidth]{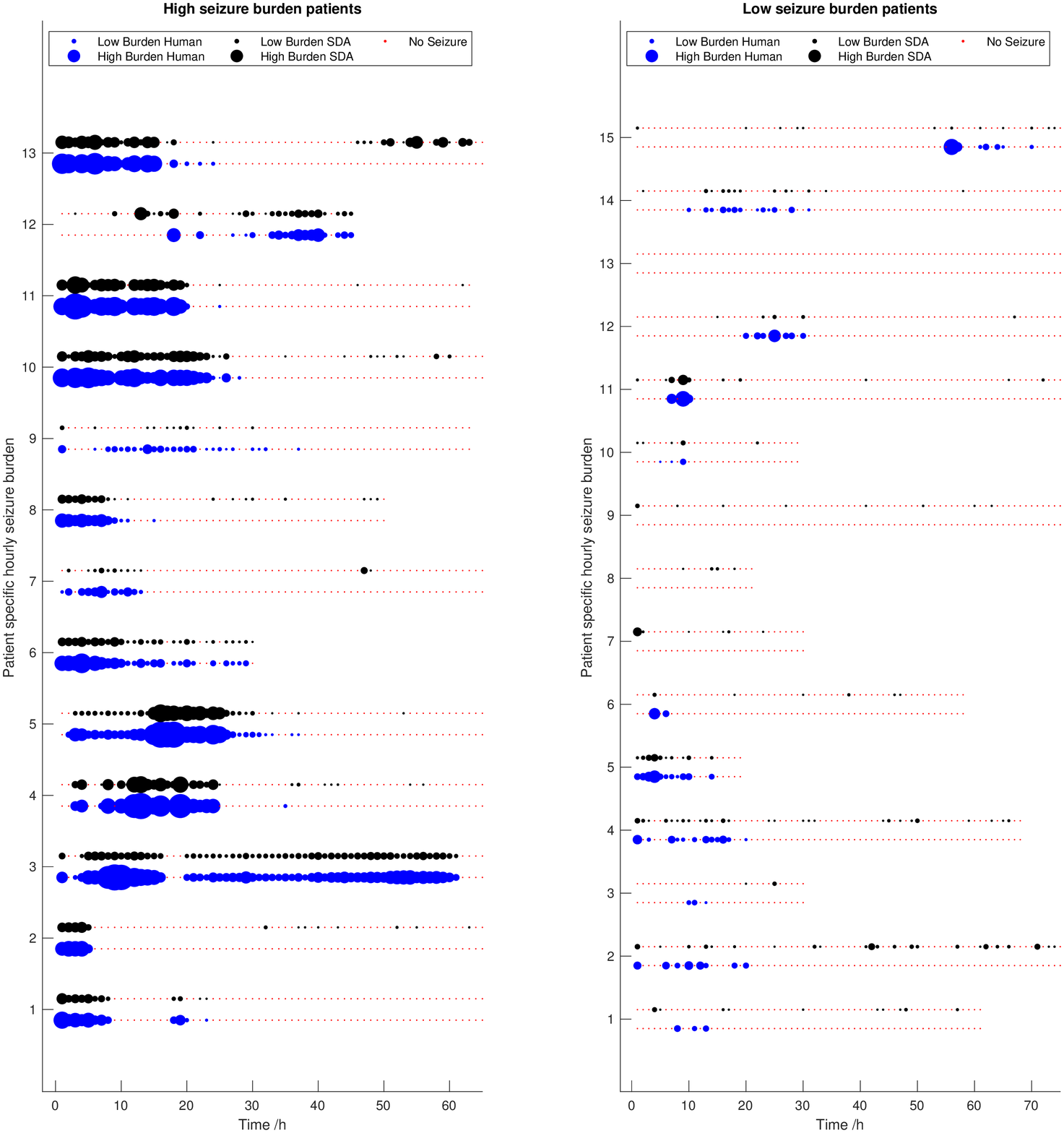}
     \caption{Hourly (short-term) seizure burden evolution over time on human consensus and SDA annotation. Each marker represents an hour segment and the marker thickness the hourly seizure burden. Time series were cut after no more events occurred for visualization purposes. Red dots stand for no seizure.}
     \label{fig:evol}
\end{figure}

To assess the clinical efficacy of the SDA, we used more clinically focused metrics (Table \ref{tab:conf_tb}). 
The key component of this assessment is evaluation of the short-term seizure burden per hour (example in Figure \ref{fig:evol}). The correlation between human consensus and SDA for short-term seizure burden per hour on concatenated data was 0.89 (95\% CI: 0.80-0.96). The correlation between human experts assessment of short-term seizure burden was 0.95 (95\% CI: 0.93-0.98).

\begin{table}[ht!]
\centering
     \caption{Agreement in clinically relevant measures between the SDA and human consensus annotation of the EEG on all neonates. POI - period of clinical interest (hourly epochs of EEG that exceed a treatment threshold for seizures), nPOI - not a POI, H - human annotation (consensus), S - algorithm annotation, HB - high seizure burden, LB - low seizure burden.}
    \begin{subtable}[h]{0.31\textwidth}
        \caption{Period of Clinical Interest (POI)}
        \begin{tabular}{c | c c}
         & POI (S) & nPOI (S) \\
         \hline
        POI (H) & 151 & 30 \\
        nPOI (H) & 103 & 716 \\
       \end{tabular}
    \label{tab:aed}
    \end{subtable}
    \begin{subtable}[h]{0.31\textwidth}
        \caption{Total Seizure Burden}
        \begin{tabular}{c | c c}
        & HB (S) & LB (S) \\
         \hline
        HB (H) & 11 & 2 \\
        LB (H) & 0 & 15 \\
       \end{tabular}
       \label{tab:prog}
    \end{subtable}
    \begin{subtable}[h]{0.31\textwidth}
                \caption{Maximum Hourly Seizure Burden}
        \begin{tabular}{c | c c}
         & HB (S) & LB (S) \\
         \hline
        HB (H) & 11 & 3 \\
        LB (H) & 0 & 14 \\        
       \end{tabular}
       \label{tab:diag}
     \end{subtable}
     \label{tab:conf_tb}
\end{table}

The hourly seizure burden is commonly used as a target to initiate treatment in clinical studies. The SDA identifies epochs that exceed a target threshold with an accuracy of 87\%  (Table \ref{tab:aed}). The sensitivity of the SDA is 83\% and the specificity is 87\% which are in a similar range to inter-observer agreement between human experts (sensitivity 89\% and specificity 97\%) for this measure.  

The sensitivity and specificity of the SDA for detecting high total seizure burden ($>45$ minutes; compared to the consensus annotation) are 85\% and 100\%, respectively (Table \ref{tab:prog}). The sensitivity and specificity of human experts are 85\% and 79\%, respectively. Finally, for detecting high maximum hourly seizure burden ($>13$ minutes per hour; compared to consensus), the sensitivity and specificity for the SDA are 79\% and 100\% (Table \ref{tab:diag}), while the sensitivity and specificity of human experts are 94\% and 92\%, respectively. Importantly, the SDA does not detect high total or hourly seizure burden in a low seizure burden neonate.

\subsection{Re-training}

Incorporating seizure and non-seizure data from the first 10 recordings in the validation set into the training data set improved AUC and Kappa on the remaining 18 neonates in the validation set (Table \ref{tab:gen_v2}). Improved sensitivity was offset by a slight reduction in specificity. These improvements were sufficient to generate an annotation that was non-inferior to the human annotation ($\Delta\kappa$ = 0.077, 95\% CI: -0.002-0.232, $n = 18$). In contrast, the original SDA, when applied to the truncated validation data set retained its inferior annotation of seizures ($\Delta\kappa$ = 0.084, 95\%CI: 0.003-0.226, $n = 18$). Re-training did not significantly change primary performance measures when evaluated within a 10-fold cross-validation on the original cohort of 79 neonates (Table \ref{tab:gen_v2}). Specificity was slightly, but significantly, reduced.


Re-training an SDA based on the Temko feature set showed slight, but not significant, improvements in performance compared to a version trained on the original cohort of 79 neonates. Compared to the proposed, retrained SDA, SDA retrained with Temko features had inferior performance with all measures except specificity and false detections per hour (Table \ref{tab:gen_v2}). The annotation of the EEG by the retrained SDA with the Temko feature set remained inferior to the human expert annotation ($\Delta\kappa$ = 0.145, 95\%CI of 0.083-0.287, $n = 18$). 


\begin{table}[ht]
\caption{\label{tab:gen_v2} The performance of the re-trained neonatal SDA. The re-trained neonatal SDA was applied to the original dataset (10-fold cross-validation) and the truncated validation set ($n=18$). These values were compared to the original SDA applied to the truncated validation set and a re-trained SDA based on the Temko feature set (Temko FS). Median (IQR) results are presented for AUC, sensitivity (Sens) and specificity (Spec). The value with 95\%CIs in brackets are presented for
cAUC, cSDR, cFD/h and cKappa (estimated on concatenated recordings). $^a$ and $^b$ denote a significant increase and decrease, respectively, between Validation and Validation (retrained), $^c$ and $^d$ denote a significant increase and decrease, respectively, between Validation (retrained) and Validation (retrained Temko FS), and $^f$ denotes a significant decrease between Training (10-fold CV) and Training (outlier detection; 10-fold CV) in Table \ref{tab:gen}. $n_t$ is the number of infants in the cohort and $n_s$ is the number of infants with consensus seizure in the cohort.}
\begin{center}
\begin{tabular}{ |c  c | c c | c |}
\hline
	& Original Training & Validation & Validation & Validation \\
	& (10-fold CV) & & (retrained) & (retrained Temko FS) \\ 
	& $n_t = 79, n_s = 39$ & $n_t = 18, n_s = 14$ & $n_t = 18, n_s = 14$ & $n_t = 18, n_s = 14 $ \\ \hline
	AUC & 0.979 (0.930-0.996) & 0.965 (0.880-0.980) & 0.976 (0.932-0.983)$^a$ & 0.953 (0.867-0.974)$^d$ \\ 
	Sens & 0.816 (0.433-0.947) & 0.519 (0.156-0.759) & 0.648 (0.311-0.837)$^a$ & 0.395 (0.057-0.645)$^d$ \\ 
	Spec & 0.991 (0.973-1)$^f$ & 0.998 (0.997-0.999) & 0.997 (0.996-0.998)$^b$ & 0.998 (0.997-0.999)$^c$ \\ \hline
    cAUC & 0.964 (0.944-0.980) & 0.964 (0.932-0.978) & 0.974 (0.953-0.984)$^a$ & 0.956 (0.912-0.971)$^d$ \\
	cSDR & 0.667 (0.515-0.778) & 0.679 (0.447-0.868) & 0.729 (0.447-0.876) & 0.600 (0.307-0.763) \\ 
	cFD/h &  1.943 (0.918-3.331) & 0.414 (0.145-0.976) & 0.332 (0.147-0.610) & 0.191 (0.092-0.303) \\
	cKappa & 0.680 (0.507-0.769) & 0.722 (0.571-0.795) & 0.735 (0.560-0.808) &  0.662 (0.484-0.726) \\ \hline 
\end{tabular}
\end{center}
\end{table}


\section{Discussion}

In this study, we have evaluated the performance of our neonatal SDA on an independent validation data set consisting of limited channel, long duration EEG recordings. The performance of the SDA was not significantly different between cross-validation on the training data set and the validation data set, using the primary generalizability measure, the AUC. While the overall performance of the SDA remained consistent, it was, nevertheless, inferior to the human expert annotation of the validation set. The inferiority of the SDA annotation was mitigated by retraining on a larger data set containing data portions of the validation data set. 

There is a limited number of validation studies performed on published neonatal SDAs \citep{gotman1997evaluation,smit2004neonatal,lawrence2009pilot,cherian2011validation,mathieson2016validation}. These studies provide seizure detection rates of 53-66\% and false detections of 0.04-2.3 per hour; values that are compatible with our SDA performance 76\% at 0.35 false detections per hour. The limitations of these studies are that 1) they do not test the generalizability of their SDAs and 2) they do not evaluate the non-inferiority of the SDA annotation compared to that of the human expert taking into account the inherent ambiguity in the annotation of multiple human experts. The algorithm with the highest congruence between training and validation sets is that of Temko \textit{et al\.}  with a seizure detection rate of 71\% at 0.25 false detection per hour on the training set and a seizure detection rate of 75\% at 0.36 false detection per hour on a validation set \citep{mathieson2016validation}. This algorithm was further shown to potentially aid clinical seizure recognition with earlier detection of seizure onset and improved temporal localization of clinical target periods \citep{mathieson2016validation}. It has recently been clinically trialled and shown to improve the recognition of hourly seizures, but not the identification on neonates with seizure, which may help optimize the timing of treatment \citep{pavel2020machine}. We have shown that our feature set is capable of outperforming the feature set of Temko \textit{et al\.} that underlies this algorithm, when trained on identical data.

Initially, the SDA provided an inferior annotation of seizures compared to human experts on the validation data set. The most obvious failing of the SDA was an excessive annotation of seizures. The SDA annotated twice as many seizures as the human experts although the majority of these seizures were multiple detections within a single human annotated seizure. Nevertheless, a considerable proportion were periods annotated by human experts as 'no seizure'. Reassuringly, visual inspection of false detections showed EEG with repetitive characteristics, activity that is commonly associate with seizure (see Figure \ref{fig:fd_fn}). 

There are two key differences between the training and validation data sets that may explain increased errors in the validation set 1) reduced electrode density and increased recording duration of EEG in the validation set, and 2) different brain injury etiologies in the two data sets. Firstly, EEG monitoring with a reduced electrode set is commonly used in the NICU as it offers less disruption for neonates, is compatible with visualizations such as the aEEG and reduces the demand of cot-side resources. While these recordings capture the majority of seizures and have electrodes located on regions that are more robust to artefact, we have shown that low density recordings result in a different annotation to standard EEG recordings and have a reduced prevalence of seizure \cite{stevenson2018channel, webb2021automated, stevenson2015interobserver, stevenson2019dataset}. This means that the manifestation of neonatal seizure may be more diffuse as the distance between seizure loci and recording electrodes will be higher. Long duration EEG also contains a wider variety of normal EEG patterns that may not be seen in short duration, higher density recordings performed early to confirm the presence of seizures. This explains why the ’no seizure’ feature values within the validation data set have a larger spread compared to the training data. Secondly, the validation set contains seizures mostly from a single aetiology (stroke) whereas the training set contained seizures from several aetiologies (HIE, stroke, meningitis) \cite{stevenson2019dataset}. Seizures in neonates with stroke may have different morphology to seizures from other aetiologies and stroke was poorly represented in the training data set. This also means that recordings in the validation set will not contain prolonged periods of isoelectric EEG and burst-suppression that are apparent in the EEG recordings of neonates with conditions such as HIE. Despite these potential limitations, the SDA had consistent, general performance between the data sets and non-inferiority to human annotation could be achieved by incorporating a subset of the validation data set into the training set.

Nevertheless, the SDA annotation provides an accurate ($\sim 90\%$) representation of key interpretations of the annotations such as identifying periods of EEG that meet treatment thresholds and definitions of high seizure burden that are associated with poor prognosis. These values may exceed current clinical practice when on-demand EEG annotation is not available, however, the risk of false identifications may outweigh increased sensitivity. A risk that is minimized for a non-inferior annotation.

Bag of features approaches to neonatal seizure detection are being superseded by deep neural networks \cite{o2020neonatal}. We and others have shown that deep convolutional neural networks provide similar performance to the proposed 'bag of features' approach when applied to our training set \cite{stevenson2019hybrid, o2020neonatal} with deep convolutional networks providing more consistent performance over several data sets \cite{o2020neonatal}. These methods will, nevertheless, face similar challenges such as lack of diversity within training data sets, subjectivity in the gold standard of human expert annotation (the training target) and performance assessment. 

A potentially controversial component of our study is the use of 10s as a minimum duration of seizure. The original choice of a 10s minimum seizure duration was always arbitrary and acknowledges that as evolution is an important component of determining the presence of seizure on an EEG recording, there must be some sort of minimum duration to judge such an evolution \citep{clancy1987exact}. This minimum seizure duration has been abolished in the latest recommendations on the Task Force for Neonatal Seizure \citep{pressler2021ilae} but its abolition is disputed \citep{shellhaas2021neonatal}. With the lack of biophysical evidence for a minimum seizure duration, it is best to turn to other forms of evidence such as studies of inter-observer agreement. These works provide the only data-driven evidence base for a minimum seizure duration and suggest that 30s may be a feasible option \citep{stevenson2015interobserver}. This, more lenient, definition of minimum seizure duration has the potential to improve the accuracy of SDAs. Furthermore, it is a simple way to achieve SDA annotations that generalize to the larger population of neonates and are non-inferior to the visual interpretation of the human expert.

\section{Conclusion}

We validated a neonatal SDA on an independent data set of neonatal EEG. We show consistent performance between training and validation data sets, but an SDA annotation that is inferior to the human expert. This annotation is, nevertheless, sufficient to provide highly accurate estimation of seizure burden and identification of clinical targets. Increasing the diversity of the training set resulted in a non-inferior SDA annotation for the left-out part of the validation data set. This raises key open questions for neonatal SDAs: 1) how much data is required and from what population of neonates should it be sampled to train neonatal SDAs that are non-inferior to the human expert, 2) is non-inferior annotation compared to human experts the ultimate target and 3) if and when validated non-inferiority is achieved, will it improve the clinical recognition of seizures and improve health outcomes for critically ill neonates.

\section{Declaration of competing interest and conflict of interest}

The authors have no competing interests to declare and no conflicts of interest.

\section{Acknowledgements}

Funding for this research was provided by the Finnish Cultural Foundation (00181077), Finnish Pediatric Foundation, the Finnish Academy (313242, 288220, 321235),  Aivosäätiö, Neuroscience Center at University of Helsinki, Helsinki University Hospital, Helsinki University Hospital Research Funds, HUS and University of Helsinki Researcher position, Sigrid Juselius Foundation and European Union’s Horizon 2020 Research and Innovation Programme (H2020-MCSA-ITN-813483 and H2020-MCSA-IF-656131). The study sponsors had no involvement in the study design, nor in the collection, analysis or interpretation of data. Additionally we acknowledge the computational resources provided by the Aalto Science-IT project.

\section{Author contribution statement}

KTT contributed to study design, developed analysis tools, performed data processing and scripting for results. PN contributed to data collection, collation and annotation of the EEG recordings. SV contributed to study design, data collection, collation and annotation of the EEG recordings. NJS contributed to study design and algorithm development. 
KTT, NJS, PN, and SV contributed to the writing and editing of the paper manuscript.


\bibliography{myrefs}

\end{document}